\documentclass[3p,times,procedia]{elsarticle}
\usepackage{ecrc}
\usepackage[bookmarks=false]{hyperref}
    \hypersetup{colorlinks,
      linkcolor=blue,
      citecolor=blue,
      urlcolor=blue}
\usepackage{amsmath}
\usepackage{booktabs} 

\volume{00}

\firstpage{1}

\journalname{Procedia Computer Science}

\runauth{Author name}

\jid{procs}

\usepackage{amssymb}

\usepackage[figuresright]{rotating}

\begin{document}
\begin{frontmatter}



\dochead{30th International Conference on Knowledge-Based and Intelligent Information \& Engineering Systems (KES 2026)}%

\title{Chunking Methods on Retrieval-Augmented Generation -- Effectiveness Evaluation Against Computational Cost and Limitations}


\author[a]{Mateusz Śmigielski} 
\author[a]{Michał Rajkowski}
\author[a]{Mateusz Zbrocki}
\author[a]{Michał Bernacki-Janson}
\author[a]{Karol Kunicki}
\author[a]{Julianna Godziszewska}
\author[a]{Maciej Piasecki}
\author[a]{Konrad Wojtasik}

\address[a]{Department of Artificial Intelligence, Faculty of Information and Communication Technology, Wrocław University of Science and Technology, Wrocław 50-370, Poland}

\begin{abstract}
Retrieval-Augmented Generation (RAG) has demonstrated significant capabilities in enhancing the performance of Large Language Models (LLMs). One of the key tasks in RAG systems is the chunking process. Traditionally, fixed-size chunking and semantic chunking have been the standard approaches. However, interest in chunking strategies has been increasing, leading to a growing number of proposed methods that often claim improved performance over these conventional techniques. Many of these approaches are tailored to specific use cases and data types, with limited evidence of their effectiveness across diverse scenarios. As a result, it remains challenging to directly compare different techniques and assess their relative strengths. To the best of our knowledge, this study is the first to systematically evaluate the effectiveness of a wide range of chunking methods and emphasize the underlying challenges of chunking strategies in RAG systems. While chunking is commonly treated as a simple preprocessing step, we show that it introduces a range of impactful and often overlooked issues.

\end{abstract}

\cortext[cor1]{Corresponding author. Tel.: +0-000-000-0000}
\end{frontmatter}

\email{author@institute.xxx}



\section{Introduction}
Retrieval Augmented Generation (RAG) has been proposed as a response to several Large Language Models (LLMs) problems like accessing, manipulating and updating knowledge or providing provenance for their decision \citep{10.5555/3495724.3496517}. 
A standard RAG pipeline initially splits documents, especially the longer ones, into smaller units called \emph{chunks}. Identifying the chunks most pertinent to a query and supplying them to the LLM -- the answer generator should allow for obtaining answers that are accurate and well aligned with the context. 
Chunks granularity is often reported to have a notable influence on both the retrieval stage and generations tasks \citep{wang2023learningfiltercontextretrievalaugmented, 10.5555/3618408.3619699}. 
Large chunks may contain irrelevant information, while small chunks may lack information, which may lead to hallucination and distract the LLM ability to extract accurate key information \citep{wang-etal-2025-document}. Additionally, \citep{wang-etal-2024-searching} demonstrate that chunking methods significantly impact retrieval performance, although their study evaluates only a limited subset of the methods. Recently, the importance of chunking has become increasingly recognized, as evidenced by emerging surveys and benchmarking studies dedicated specifically to document segmentation strategies. Zhou et al. \citep{zhou2026chunkthenembedcomprehensivetaxonomyevaluation} highlight the diversity of document chunking methods in dense retrieval and emphasize the need for a systematic taxonomy to understand their design space. They categorize existing strategies based on segmentation approaches and embedding paradigms, providing a framework for comparing and evaluating chunking techniques. Renyi Qu et al. \citep{qu-etal-2025-semantic}  evaluate the effectiveness of semantic chunking and demonstrate that its computational costs are not justified by consistent performance gains. The study compares semantic chunking with simpler fixed-length baselines and analyzes trade-offs between retrieval quality and efficiency. The results indicate that while semantic chunking can improve contextual coherence and, in certain scenarios, retrieval performance, the gains are not always proportional to the increased computational cost. This suggests that the adoption of advanced chunking techniques should be guided by practical constraints and system requirements, especially in large-scale or latency-sensitive applications.
The relevance of chunking has also been explicitly acknowledged in medical applications, where retrieval errors may directly affect clinical reasoning. In a recent study, Gomez-Cabello et al. \citep{bioengineering12111194} conduct a comparative evaluation of chunking strategies within RAG systems designed for clinical decision support. Their analysis demonstrates that chunking methodology significantly influences RAG-LLM performance for clinical decision support. However, it should be noted that the scope of chunking strategies considered in that study was relatively limited and did not cover the full spectrum of existing segmentation methods. 

Our contribution in this study is threefold. First, we present an evaluation framework designed to enable the systematic comparison of a broad range of chunking methods across diverse tasks and dataset types. Second, we identify several practical and methodological challenges, referred to as efficiency limitations, including high chunking time and implementation fragility. Third, we show that more computationally expensive chunking methods do not yield meaningful effectiveness improvements while introducing substantially higher computational overhead. To facilitate systematic comparison and reproducible analysis, we provide a unified evaluation framework available at \url{https://github.com/ApriiM/Chunking-Research}.

The remainder of this paper is structured as follows. First, we provide a review of existing document chunking methods, outlining their key characteristics and differences. We then describe the chunk preparation pipeline, the dataset selection process and the evaluation methodology employed in our study. This is followed by a presentation and analysis of the experimental results, highlighting performance differences, limitations, and inconsistencies across methods. Finally, we summarize our findings and discuss potential directions for future research.

\section{Related works}
\subsection{Theoretical Background}
\label{sec:theory}

Text segmentation, often referred to as chunking, is a preprocessing step that transforms a continuous text stream into discrete text segments. 

Formally, let us define a document $D$ as a sequence of atomic units (tokens) $D = \langle t_1, t_2, \dots, t_N\rangle$. The goal of the chunking function $f_{\theta}$ is to map $D$ into a set of segments $\mathcal{C} = \{c_1, c_2, \dots, c_m\}$, such that:
$c_i = \langle t_{start_i}, \dots, t_{end_i}\rangle$.


While strictly disjoint segmentation is possible, overlapping segmentation is often employed to preserve boundary contexts.

\noindent \emph{The Granularity-Context Trade-off.}
The fundamental theoretical challenge is maximizing the probability of generating the correct answer $a$ given a query $q$. This introduces a trade-off between \textit{granularity} and \textit{context preservation}.

\noindent \emph{Chunking in the RAG Pipeline.}
In the vector space model, chunking defines the granularity of the search index. Let $\mathcal{E}$ be an embedding function (e.g., BERT-based encoder) mapping a text segment to a vector $v \in \mathbb{R}^d$ \citep{karpukhin2020dense}. The retrieval process identifies a subset $\mathcal{C}_{retrieved} \subset \mathcal{C}$ by minimizing the distance metric:

\begin{equation}
    \mathcal{C}_{retrieved} = \underset{c \in \mathcal{C}}{\mathrm{arg\,topk}} \left( \frac{\mathcal{E}(q) \cdot \mathcal{E}(c)}{\|\mathcal{E}(q)\| \|\mathcal{E}(c)\|} \right)
\end{equation}

Ineffective segmentation leads to a mismatch in the embedding space, where the vector representation $\mathcal{E}(c)$ is dominated by irrelevant content, causing semantic misalignment with the query $\mathcal{E}(q)$ \citep{reimers2019sentence}.

\subsection{Chunking methods}
The foundational strategies for text segmentation rely on surface-level features such as character count or punctuation, ignoring semantic content. \emph{Fixed-size Chunking} divides text into segments of predetermined length, often with overlapping windows to reduce mid-sentence splits, but it frequently fragments semantic dependencies \citep{langchain}. \emph{Sentence/Paragraph Splitting} preserves linguistic units like sentences or paragraphs, though chunk sizes vary widely, which can limit context for embedding models. \emph{Recursive Character Splitting} uses a hierarchy of separators (paragraphs, lines, spaces) to iteratively split oversized segments, preserving semantic coherence when possible. \emph{Entropy-Optimized Dynamic Text Segmentation (EDTS)} determines chunk boundaries by analyzing lexical homogeneity via information entropy \citep{wang-edts-2025}. \emph{Semantic Chunking} detects topic transitions by monitoring similarity drops between consecutive sentences, aligning chunks with thematic coherence but incurring high computational cost and threshold selection challenges \citep{kamradt2024semantic}.

LLM-based chunkers leverage language models to produce semantically coherent text segments. \emph{LumberChunker} \citep{duarte-etal-2024-lumberchunker} splits documents into paragraphs and aggregates them into groups passed to an LLM, which identifies points of semantic shift to finalize chunks. Evaluated on GutenQA, it outperforms Semantic, Recursive, and Proposition-Level Chunking, but incurs high computational cost and is mainly suited for narrative texts. \emph{Propositions} \citep{chen-etal-2024-dense} move from chunk-level to finer-grained atomic units encapsulating distinct factoids, automatically generated via a propositionizer and indexed for retrieval, related to prior notions of factoids and information nuggets \citep{Pavlu2012IRSE, pradeep2025greatnuggetrecallautomating}. \emph{Pseudo-Instruction for Document Chunking} \citep{wang-etal-2025-document} groups consecutive sentences based on semantic similarity to a document summary, forming alternating segments of highly related versus less related content. \emph{HiChunk} \citep{lu2026hichunk} detects Global Chunk Points with a fine-tuned LLM, assigning hierarchical levels and iteratively processing fragments to handle long documents; a fixed-size post-chunking step and Auto-Merge Retrieval improve uniformity and query-aligned retrieval. \emph{Logits-Guided Multi-Granular Chunker (LGMGC)} combines Logits-Guided Chunking, which uses LLM token probabilities to detect semantic endpoints, with Multi-Granular Chunking, recursively subdividing parent chunks into smaller levels, ranking all chunks by similarity to the query for final retrieval. \emph{Meta-Chunking} framework \citep{zhao2025metachunkinglearningtextsegmentation} introduces Perplexity Chunking and Margin Sampling Chunking to enhance the semantic integrity and contextual coherence of chunks. Additionally, it integrates meta-chunking with dynamic merging and employs a global information compensation mechanism via hierarchical summarization and multi-stage chunk rewriting to repair semantic discontinuities caused by segmentation.

Adaptive approaches avoid committing to a single segmentation strategy. \emph{Mix-of-Granularity (MoG)} \citep{zhong-etal-2025-mix} pre-segments documents into multiple granularity levels, and a query-conditioned routing module selects which level to use for retrieval, aligning chunk length and specificity with query needs. \emph{Mixtures of Chunking Learners (MoC)} \citep{zhao-etal-2025-moc} combines multiple chunking models to generate alternative segmentations, after which a router selects a meta-chunker model based on chunk-quality metrics. The chosen SLM-based meta-chunker is trained to predict only structured chunk boundaries, masking central content during training, which allows efficient reconstruction from the original document while reducing computational overhead. This design highlights the complementary strengths of different chunking strategies and improves semantic alignment with queries.

Clustering-based chunking treats segmentation as a grouping task, identifying semantically similar units rather than cutting text arbitrarily. \emph{Sequential Hierarchical Agglomerative Chunking} \citep{qu-etal-2025-semantic} merges adjacent sentences based on semantic similarity while enforcing a strict structural constraint, ensuring narrative continuity without tuning positional weights or predefining the number of chunks. \emph{Max-Min Semantic Chunking} \citep{kiss2025maxmin} uses a greedy strategy that sequentially adds sentences to a chunk only if their maximum similarity exceeds an adaptive threshold derived from the chunk’s minimum coherence, dynamically controlling chunk growth. 

Recent structural chunkers leverage document layout and element types (titles, tables, figures) detected by vision models or object detectors to guide segmentation.
\emph{AutoChunker} converts documents into Markdown, splits into sentences, and uses an LLM to merge or discard sentences, producing hierarchical chunks based on semantic structure \citep{jain-etal-2025-autochunker}. \emph{cAST} segments code using Abstract Syntax Trees, producing chunks aligned with syntactic boundaries \citep{zhang-etal-2025-cast}. \emph{S2 Chunking} builds a weighted graph over elements using spatial proximity and semantic similarity, then applies spectral clustering to form chunks, enforcing a maximum token length per chunk \citep{verma2025s2chunkinghybridframework}.

An interesting alternative approach for chunked document contextualization is \emph{Late Chunking} \citep{günther2025latechunkingcontextualchunk}. Instead of adding extra context text to each chunk (e.g., summaries or surrounding passages), Late Chunking adds context at the embedding level. A long-context embedding model first encodes tokens from the entire document, producing token-level representations. Then, chunk boundaries are computed on the original text, and the token embeddings within each chunk are pooled into a single vector. 

Additionally, some classical chunking approaches from earlier NLP research have been largely overlooked in the context of RAG. For instance, \emph{TextTiling} \citep{hearst1997texttiling} was an early method designed to detect topical shifts in texts based on lexical cohesion. Despite their historical relevance, these techniques are rarely included in modern evaluations of chunking strategies for RAG. As a result, it remains unclear whether such linguistically motivated approaches could complement or even outperform certain contemporary embedding- or model-based chunking methods.

\section{Methodology}
To comprehensively cover different approaches to chunking, we selected a diverse set of chunking methods representing classical (Fixed Size, TextTiling), semantic (Recursive Semantic), clustering-based (Sequential HAC Chunker, Max Min Chunker), graph-based (GraphSeg), and LLM-based (Lumberchunker) approaches. Default hyperparameters were taken from the original publications introducing the methods or, when not explicitly specified, from the corresponding official code repositories, with the exception of Lumberchunker, where we used GPT-OSS-20B. Additionally, we modified the TextTiling implementation so that chunk boundaries were aligned to the nearest sentence rather than the nearest paragraph. Preliminary experiments indicated that paragraph-level normalization frequently resulted in no effective segmentation at all, often returning the original documents without introducing meaningful chunk boundaries across nearly all evaluated datasets. In contrast, sentence-level alignment yielded substantially more coherent and stable chunks. Since Fixed-Size Chunking is a simple heuristic rather than a formally grounded method, we used a chunk size of 512 (i.e.\ quite common choice) with an overlap of 50. Consequently, chunk sizes were not normalized across methods, as we wanted each method to freely adapt its chunking strategy according to its design assumptions. The resulting configurations were then used to segment documents across the selected datasets. Datasets were selected to fulfill requirements and cover a wide range of domains to enable reliable evaluation of the chunking methods. Additionally, due to the lack of sufficiently challenging datasets, we merged each of the three selected datasets into a single massive document to create stress-test conditions for the evaluated methods (\ref{app:datasets}). In order to cope with the complexity of the problem, we had to delimit the scope of our experimental study, imposing a 48-hour time limit for each chunking process, as we considered any method exceeding this duration to introduce excessive computational overhead. At this stage, we also propose a principled evaluation framework for the qualitative analysis of the generated chunks that we include in our code repository. After preparing the chunks, we designed two experiments corresponding to two different stages of the RAG pipeline.

\subsection{Evidence retrieval}
The first experiment evaluated evidence retrieval, focusing on the effectiveness of chunking methods in retrieving relevant chunks for a given query. A chunk was considered relevant if it overlapped with the span of the extractive answer. In cases where no extractive answer was available, all chunks originating from the relevant document were treated as relevant. To ensure a fair comparison across methods, we evaluated the number of relevant chunks included in the top-k retrieved results using the unified Polish Information Retrieval Benchmark (PIRB) \citep{dadas-etal-2024-pirb} with accuracy and recall as the evaluation metrics. During the retrieval evaluation phase, the bge-m3 model was employed as the primary retriever, while bge-reranker-v2-m3 served as the reranking model.

\subsection{End-to-end RAG Answer Generation}
The second experiment evaluated the impact of chunking methods on the end-to-end RAG system and the quality of LLM-generated answers. We employed the GPT-OSS-20B model as the generative model. For each query, the top-5 retrieved chunks were provided as contextual input with a restriction to 4,000 tokens. The generated responses were then evaluated against ground-truth answers using an LLM-as-a-judge approach \cite{10.5555/3666122.3668142}, with GPT-OSS-20B serving as the judge model.
The evaluation was conducted using a five-point Likert scale, where scores ranged from 1 (poor answer quality and relevance) to 5 (highly accurate and relevant answer). The experiments were conducted on a subset of the original datasets due to missing answers in some datasets. Additionally, the TextTiling method was excluded from the comparison as it relied on a modified implementation that was not directly comparable to the other chunking approaches under a unified experimental setup.

\section{Experiments and Results}
For each chunker, text segmentation was performed on all datasets included in the study. The segmentation process was conducted in parallel on 20 nodes with the following hardware specifications: CPU: 2x Intel Xeon Platinum 8462Y (32 cores, 2.8 GHz), GPU: 4x Nvidia H100 96GB, 16896 CUDA cores, 1980 MHz, RAM: 1006 GB 4800 MT/s ECC DDR5. So, approximately, each was run on 1/4 of a node. 

\begin{table}[h]
\centering
\caption{Accuracy@5 for evidence retrieval. T - chunking procedure exceeded the allotted execution time; S - The underlying spaCy library triggered an error due to excessive temporary memory requirements.}
\label{tab:accuracy}
\begin{tabular*}{\textwidth}{@{\extracolsep{\fill}}lllllllll@{}}
\toprule
Dataset & DenseX & \shortstack{Recursive\\semantic} & Fixed-size & \shortstack{Sequential\\HAC} & Text Tiling & Max-min & GraphSeg & Lumberchunker \\
\colrule
GutenQA & T & \textbf{92.42} & 87.86 & 71.45 & 82.12 & 82.72 & 88.53 & T\\
\shortstack{GutenQA\\merged} & T & \textbf{89.70} & 87.49 & 65.92 & 77.66 & 78.23 & S & T\\
LiteraryQA & T & \textbf{87.59} & 85.91 & 75.73 & 82.81 & 83.40 & S & T\\
NQ & 97.33 & \textbf{99.00} & 98.00 & \textbf{99.00} & \textbf{99.00} & \textbf{99.00} & 98.67 & T\\
NovelQA & T & \textbf{91.61} & 90.10 & 82.90 & 88.07 & 88.53 & S & T\\
PoQuAD & 25.80 & \textbf{97.68} & 96.39 & 88.74 & 96.36 & 93.58 & 97.45 & 97.05\\
\shortstack{PoQuAD\\merged} & T & 93.11 & 94.67 & 88.86 & 91.74 & 93.62 & S & \textbf{97.48}\\
Qasper & 49.55 & 49.97 & 51.00 & 51.27 & 50.52 & \textbf{51.48} & 49.62 & 50.10\\
SQuAD & 85.13 & 96.87 & 96.24 & 94.60 & 95.81 & 96.24 & \textbf{97.34} & 97.10\\
TriviaQA & 79.50 & \textbf{91.80} & 87.30 & 78.30 & 85.10 & 86.30 & 89.50 & T\\
\shortstack{TriviaQA\\merged} & 77.30 & \textbf{93.20} & 89.80 & 84.20 & 85.40 & 90.10 & S & T\\
\midrule
Average & 69.10 & \textbf{89.36} & 87.71 & 80.09 & 84.96 & 85.75 & 86.85 & 85.44\\
\botrule
\end{tabular*}
\end{table}

\begin{table}[h]
\centering
\caption{Recall@10 for evidence retrieval. T - chunking procedure exceeded the allotted execution time; S - The underlying spaCy library triggered an error due to excessive temporary memory requirements.;}
\label{tab:recall}
\begin{tabular*}{\textwidth}{@{\extracolsep{\fill}}lllllllll@{}}
\toprule
Dataset & DenseX & \shortstack{Recursive\\\\semantic} & Fixed-size & \shortstack{Sequential\\\\HAC} & Text Tiling & Max-min & GraphSeg & Lumberchunker\\
\colrule
GutenQA & T & \textbf{67.62} & 39.65 & 11.31 & 23.48 & 26.58 & 47.74 & T\\
\shortstack{GutenQA\\merged} & T & \textbf{53.72} & 36.57 & 10.92 & 20.34 & 25.50 & S & T\\
LiteraryQA & T & \textbf{7.18} & 4.53 & 0.62 & 2.02 & 1.81 & S & T\\
NQ & 4.28 & \textbf{30.18} & 16.69 & 5.11 & 13.08 & 10.34 & 18.02 & T\\
NovelQA & T & \textbf{0.92} & 0.44 & 0.07 & 0.19 & 0.22 & S & T\\
PoQuAD & 24.01 & \textbf{98.31} & 96.26 & 89.86 & 96.95 & 94.69 & 98.01 & 97.66\\
\shortstack{PoQuAD\\merged} & T & 82.85 & 82.81 & 79.25 & 82.02 & 83.55 & S & \textbf{87.05}\\
Qasper & 11.71 & 46.14 & 34.37 & 18.25 & 26.98 & 35.66 & \textbf{48.95} & 40.81\\
SQuAD & 66.28 & 85.81 & 81.81 & 79.84 & 81.79 & 84.24 & 85.98 & \textbf{87.14}\\
TriviaQA & 38.42 & \textbf{78.66} & 64.90 & 49.95 & 61.23 & 62.05 & 71.78 & T\\
\shortstack{TriviaQA\\merged} & 19.90 & \textbf{40.52} & 34.18 & 27.19 & 30.27 & 33.82 & S & T\\
\midrule
Average & 27.43 & 53.81 & 44.75 & 33.85 & 39.85 & 41.68 & 61.75 & \textbf{78.16}\\
\botrule
\end{tabular*}
\end{table}

\begin{table}[h]
\centering
\caption{Chunking time per dataset and method}
\label{tab:time}
\begin{tabular*}{\textwidth}{@{\extracolsep{\fill}}lllllllll@{}}
\toprule
Dataset & DenseX & \shortstack{Recursive\\\\semantic} & Fixed-size & \shortstack{Sequential\\\\HAC} & Text Tiling & Max-min & GraphSeg & Lumberchunker \\
\colrule
GutenQA & - & 10.99m & \textless1s & 8.72m & 2.22m & 9.18m & 1.13h & -\\
\shortstack{GutenQA\\merged} & - & 11.89m & \textless1s & 3.27m & 6.37m & 4.09m & - & -\\
LiteraryQA & - & 9.91m & \textless1s & 3.19m & 2.69m & 4.16m & - & -\\
NQ & 30.79h & 2.98m & \textless1s & 2.58m & 36s & 2.84m & 11.68h & -\\
NovelQA & - & 11.02m & \textless1s & 3.18m & 2.66m & 4.03m & - & -\\
PoQuAD & 3.05h & 43s & \textless1s & 17s & 23s & 17s & 47s & 9.94h\\
\shortstack{PoQuAD\\merged} & - & 22s & \textless1s & 4s & 24s & 5s & - & 4.27h\\
Qasper & 34.34m & 15s & \textless1s & 5s & 7s & 4s & 17s & 1.90h\\
SQuAD & 5.47h & 39s & \textless1s & 7s & 16s & 9s & 32.55m & 17.36h\\
TriviaQA & 25.39h & 2.60m & \textless1s & 54s & 38s & 1.10m & 5.17h & -\\
\shortstack{TriviaQA\\merged} & 25.03h & 2.51m & \textless1s & 39s & 48s & 50s & - & -\\
\midrule
Average & 15.05h & 4.90m & \textbf{\textless1s} & 2.10m & 1.56m & 2.44m & 3.09h & 8.37h\\
\botrule
\end{tabular*}
\end{table}

\begin{figure}[t]
\vspace*{4pt}
\centering
\label{fig:number_of_chunks}
\includegraphics[width=0.8\columnwidth]{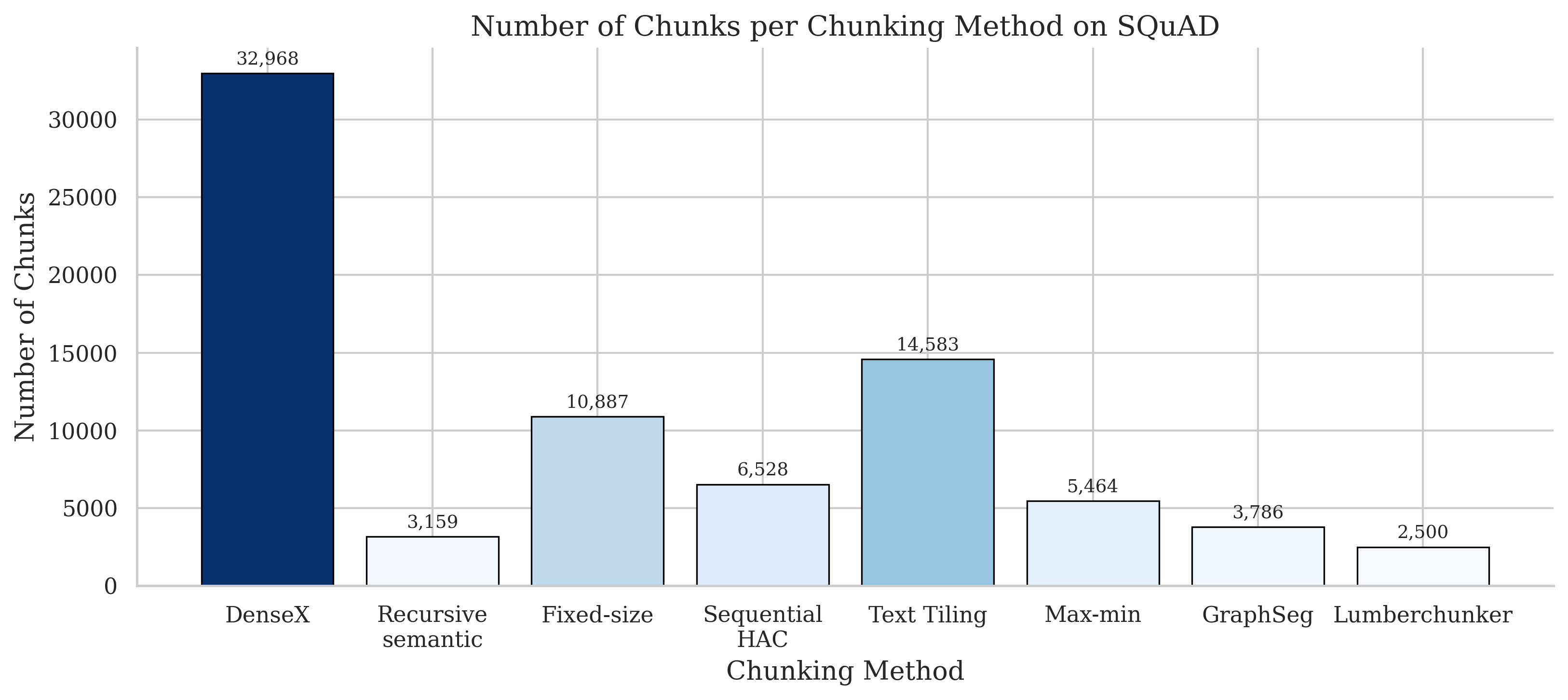}
\caption{Number of chunks per chunking method on SQuAD dataset.}
\label{fig:your_label}
\end{figure}

\begin{table}[h]
\centering
\caption{LLM-as-a-judge score in answer generation task}
\label{tab:RAGEval}
\begin{tabular*}{\textwidth}{@{\extracolsep{\fill}}llllllll@{}}
\toprule
Dataset & DenseX & \shortstack{Recursive\\\\semantic} & Fixed-size & \shortstack{Sequential\\\\HAC} & Max-min & GraphSeg & Lumberchunker\\\\
\colrule
GutenQA & - & \textbf{3.65} & 3.30 & 2.50 & 3.05 & 3.41 & -\\
\shortstack{GutenQA\\merged} & - & \textbf{3.59} & 3.25 & 2.48 & 3.02 & - & -\\
LiteraryQA & - & \textbf{2.10} & 1.93 & 1.45 & 1.65 & - & -\\
PoQuAD & 3.77 & 4.24 & \textbf{4.25} & 3.84 & 4.07 & 4.24 & 4.21\\
\shortstack{PoQuAD\\merged} & 3.09 & 4.09 & 4.18 & 3.87 & 4.08 & - & \textbf{4.28}\\
SQuAD & 4.44 & 4.56 & 4.57 & 4.44 & 4.53 & \textbf{4.58} & 4.57\\
TriviaQA & 3.99 & \textbf{4.37} & 4.32 & 4.09 & 4.22 & 4.30 & -\\
\shortstack{TriviaQA\\merged} & 3.84 & \textbf{4.32} & 4.26 & 4.16 & 4.22 & - & -\\
\midrule
Average & 3.83 & 3.86 & 3.76 & 3.35 & 3.61 & 4.13 & \textbf{4.35}\\
\botrule
\end{tabular*}
\end{table}

Tab.~\ref{tab:accuracy} and \ref{tab:recall} present the results of the first experiment -- evidence retrieval, respectively: Accuracy@5 and Recall@10, across a total of 88 experimental configurations. The best-performing results for each dataset are in bold. 

Across experiments, a substantial number of configurations did not produce usable outputs. We identified them as practical and methodological issues. With the exception of TextTiling and HAC Semantic Chunker, our experiments utilized the original implementations provided by the authors of the respective chunking methods. Consequently, any issues related to code optimization, execution bottlenecks, or structural failures reflect the current state of these specific implementations. When an error or failure occurred, the corresponding configuration was excluded from further downstream evaluation, as resolving third-party implementation inefficiencies falls outside the scope of this comparative study. The encountered failures are denoted in the tables using the following markers: The marker “T” indicates severe scalability and efficiency limitations; in many cases, the chunking process required more than 48 hours to complete. The marker "S" further reveals implementation fragility, where spaCy-related errors prevented the processing of long documents exceeding one million characters. 

An analysis of these failures reveals significant robustness issues across several approaches. GraphSeg performs well when execution succeeds, showing competitive retrieval quality across both metrics, but its applicability is limited by memory-related failures in the underlying implementation. Lumberchunker is highly inconsistent, occasionally achieving excellent retrieval performance, but failing on many datasets due to time constraints, which significantly reduces its overall robustness. These limitations are largely attributable to its reliance on an underlying LLM, whose behavior can lead to overly conservative or inconsistent segmentation decisions. DenseX exhibited lower performance across most experimental settings, frequently failing entirely or producing low-quality outputs when execution was possible. Overall, only Recursive Semantic, Fixed-size, Max-min, Text Tiling and Sequential HAC consistently provide sufficient robustness for full evaluation coverage, with the first two offering the best trade-off between stability and retrieval effectiveness.

Table \ref{tab:time} details the chunking execution times across the experiments. The results clearly establish Fixed-size chunking as the most computationally efficient method, significantly outperforming all other approaches. Conversely, DenseX emerged as the slowest method by a substantial margin. The remaining techniques fall within an intermediate range: Sequential HAC, Max-min, Text Tiling, and Recursive Semantic exhibited comparable, moderate runtimes. GraphSeg and Lumberchunker, however, required the longest execution times after DenseX, indicating considerably higher computational overhead.

Figure \ref{fig:number_of_chunks} shows the number of chunks per chunking method on SQuAD dataset. Methods with fewer but more coherent chunks (e.g., GraphSeg and Lumberchunker) achieve the strongest Recall@10 and Accuracy@5, despite producing far fewer chunks than over-segmented approaches like DenseX or TextTiling. Higher chunk counts in simpler baselines do not consistently translate into better results, indicating a limited benefit from increased segmentation density alone. Overall, chunk quality and structural coherence are more important than chunk quantity for both Recall@10 and Accuracy@5.

In second experiment, the end-to-end RAG answer generation results in Tab.~\ref{tab:RAGEval} show that Lumberchunker achieves the highest average score, but this is based only on a limited number of successful runs, which makes the estimate less reliable and prevents strong conclusions about its overall advantage. Excluding this limitation, GraphSeg, Recursive Semantic and DenseX consistently perform among the best methods. Fixed-size chunking remains a solid baseline that is competitive with more complex approaches, while Max-min performs below these methods but still outperforms Sequential HAC, which is the weakest method overall. Overall, the findings indicate that although advanced methods can provide marginal gains, simpler approaches remain highly competitive, and any apparent advantage of LLM-based chunking should be interpreted cautiously due to incomplete coverage.

\section{Conclusion and Future Work}
Our experiments show that chunking in RAG systems is a much more difficult and fragile problem than is often assumed. Many methods fail in practice due to timeouts, memory issues, or poor scalability, and only a subset (Recursive Semantic, Fixed-size, Sequential HAC, Max-min) consistently completes processing across datasets. Results also vary strongly depending on document type and implementation constraints, and the execution time differences between methods are substantial. This indicates that chunking performance cannot be assessed without considering robustness and efficiency, not only retrieval metrics. Future work should focus on further systematic and fair evaluation of chunking methods across diverse datasets and document types as well as extending evaluations to multiple embedding models and reranking strategies to assess generalisability across retrieval components. Comparisons in literature are often limited and potentially biased toward specific scenarios. A unified taxonomy that includes robustness, scalability, and computational cost is needed. Overall, chunking should be treated as a core research problem in RAG systems rather than a simple preprocessing step. Only such a comprehensive and standardized comparison will make it possible to answer whether chunking methods have a meaningful impact on RAG quality, or whether observed differences are mainly driven by dataset selection and implementation effects.

\section*{Acknowledgements}
Financed by: CLARIN-PL project financed as part of the investment: "CLARIN ERIC -- European Research Infrastructure Consortium: Common Language Resources and Technology Infrastructure (2024-2026), funded by the Polish Ministry of Science and
Higher Education (
2024/WK/01).

\appendix
\section{Datasets}
\label{app:datasets}
The goal was to ensure a meaningful and fair evaluation of chunking strategies across diverse domains and document structures. First, document length plays a critical role in chunking evaluation. Chunking is inherently irrelevant for short texts, as there is little to no need for segmentation. Consequently, many widely used QA datasets were excluded or deemphasized because their documents are too short to meaningfully benefit from chunking. Second, most existing datasets do not provide explicit annotations of \emph{golden chunks}, i.e., text segments that optimally contain the information needed to answer a query. To address this limitation, we selected datasets that mostly include answer annotations grounded in the source documents (e.g., answer spans, supporting passages, or evidence labels). These annotations allow us to approximate gold chunks indirectly by identifying regions of the text that contain relevant information.
The datasets used in our study span multiple domains and levels of complexity:
\begin{itemize}
    \item \emph{Narrative and literary datasets:} \\
    \textit{GutenQA} \citep{duarte-etal-2024-lumberchunker}, \textit{LiteraryQA} \citep{bonomo-etal-2025-literaryqa}, and \textit{NovelQA} \citep{wang2025novelqabenchmarkingquestionanswering} provide long-form narrative texts, often derived from books (e.g., Project Gutenberg). These datasets feature document lengths ranging from several thousand characters to entire novels. They are particularly suitable for evaluating chunking in story-based QA, where relevant information may be distributed across distant parts of the text.

    \item \emph{Scientific and structured documents:} \\
    \textit{Qasper} \citep{dasigi2021datasetinformationseekingquestionsanswers} focuses on scientific papers with structured sections (e.g., headings and paragraphs). It includes explicit evidence annotations, making it valuable for evaluating chunking methods that leverage document structure (e.g., section-aware or hierarchical chunking).

    \item \emph{Open-domain and multi-domain QA datasets:} \\
    \textit{TriviaQA} \citep{joshi-etal-2017-triviaqa}, \textit{SQuAD} \citep{rajpurkar2016squad100000questionsmachine}, \textit{PoQuAD} \citep{10.1145/3587259.3627548} and \textit{Natural Questions} \citep{kwiatkowski-etal-2019-natural} 
    a wide range of topics and 
    sources.
\end{itemize}

In summary, dataset selection for chunking evaluation requires careful balancing of three factors: (1) sufficiently long documents, (2) availability of answer-grounded annotations, and (3) diversity of domains and structures. Despite the lack of explicit \emph{golden chunks} in several datasets, the chosen datasets provide a practical foundation for assessing chunking strategies through their impact on retrieval performance. The selected datasets statistics are presented in Table \ref{tab:datasets}.

\begin{table}[h]
\centering
\caption{Datasets statistics.}
\label{tab:datasets}
\begin{tabular*}{\textwidth}{@{\extracolsep{\fill}}lllll@{}}
\toprule
Dataset & Documents & Min document length & Max document length & Avg document length\\
\colrule
GutenQA & 36,917 & 4 & 32,767 & 1,814\\
GutenQA merged & 1 & 67,058,241 & 67,058,241 & 67,058,241\\
LiteraryQA & 138 & 1,459 & 1,833,987 & 411,471\\
Natural Questions & 300 & 82 & 215,119 & 45,634\\
NovelQA & 60 & 322,441 & 6,839,619 & 1,007,786\\
PoQuAd & 1,449 & 501 & 3,860 & 922\\
Qasper & 416 & 314 & 2,003 & 1,014\\
SQuAD & 100 & 3,715 & 92,637 & 29,971\\
TriviaQA & 1,000 & 113 & 430,874 & 14,239\\
TriviaQA merged & 1 & 14,241,172 & 14,241,172 & 14,241,172\\
\botrule
\end{tabular*}
\end{table}





\bibliography{bibliography}
\bibliographystyle{elsarticle-harv}

\end{document}